\documentclass[conference]{IEEEtran}

\ifCLASSOPTIONcompsoc
\else
\fi
\ifCLASSINFOpdf
\else
\fi

\usepackage{amssymb}
\usepackage{amsmath}
\usepackage{graphicx}
\usepackage{epsfig}
\usepackage{multirow}
\usepackage{setspace}
\usepackage{bbm}
\usepackage{type1cm}
\usepackage{lettrine}

\usepackage{algpseudocode}
\usepackage{algorithm}

\usepackage{url}
\usepackage{breqn}

\usepackage[switch]{lineno}

\long\def\symbolfootnote[#1]#2{\begingroup%
\def\thefootnote{\fnsymbol{footnote}}\footnote[#1]{#2}\endgroup} 

\newcommand\Mark[1]{\textsuperscript{#1}}

\begin{document}
%
\title{Scene Text Detection via Holistic, Multi-Channel Prediction}
%
%
%
%

\author{\IEEEauthorblockN{Cong Yao\Mark{1,2}, Xiang~Bai\Mark{1}, Nong Sang\Mark{1}, Xinyu Zhou\Mark{2}, Shuchang Zhou\Mark{2}, Zhimin Cao\Mark{2}}
\IEEEauthorblockA{\Mark{1} Huazhong University of Science and Technology (HUST). Wuhan, China.\\
\Mark{2} Megvii Inc. Beijing, China.\\
Email: yaocong2010@gmail.com, \{xbai, nsang\}@hust.edu.cn, \{zxy, zsc, czm\}@megvii.com}}

%
%

\markboth{ }%
{Yao~\MakeLowercase{\textit{et al.}}: Scene Text Detection via Holistic, Multi-Channel Prediction}
%


\IEEEcompsoctitleabstractindextext{%
\begin{abstract}
Recently, scene text detection has become an active research topic in computer vision and document analysis, because of its great importance and significant challenge. However, vast majority of the existing methods detect text within local regions, typically through extracting character, word or line level candidates followed by candidate aggregation and false positive elimination, which potentially exclude the effect of wide-scope and long-range contextual cues in the scene. To take full advantage of the rich information available in the whole natural image, we propose to localize text in a holistic manner, by casting scene text detection as a semantic segmentation problem. The proposed algorithm directly runs on full images and produces global, pixel-wise prediction maps, in which detections are subsequently formed. To better make use of the properties of text, three types of information regarding text region, individual characters and their relationship are estimated, with a single Fully Convolutional Network (FCN) model. With such predictions of text properties, the proposed algorithm can simultaneously handle horizontal, multi-oriented and curved text in real-world natural images. The experiments on standard benchmarks, including ICDAR 2013, ICDAR 2015 and MSRA-TD500, demonstrate that the proposed algorithm substantially outperforms previous state-of-the-art approaches. Moreover, we report the first baseline result on the recently-released, large-scale dataset COCO-Text.
\end{abstract}

}

\maketitle

\IEEEdisplaynotcompsoctitleabstractindextext

%
\IEEEpeerreviewmaketitle

\vspace{-4mm}
\section{Introduction}

Textual information in natural scenes can be very valuable and beneficial in a variety of real-world applications, ranging from image search~\cite{Ref:Tsai2011}, human-computer interaction~\cite{Ref:Kisacanin2005}, to criminal investigation~\cite{Ref:TRAIT2015} and assistance for the blind~\cite{Ref:Yi2012}. In the past few years, scene text detection and recognition have received a lot of attention from both the computer vision community and document analysis community, and numerous inspiring ideas and methods~\cite{Ref:Chen2004, Ref:Epshtein2010, Ref:Neumann2010, Ref:Wang2011, Ref:Neumann2012, Ref:Mishra2012, Ref:Novikova2012, Ref:Bissacco2013, Ref:Yin2014, Ref:Jaderberg2015B} have been proposed to tackle these problems.

However, localizing and reading text in uncontrolled environments (i.e., in the wild) are still overwhelmingly challenging, due to a number of factors, such as variabilities in text appearance, layout, font, language and style, as well as interferences from background clutter, noise, blur, occlusion, and non-uniform illumination. In this paper, we focus on the problem of scene text detection, which aims at predicting the presence of text, and if any, estimating the position and extent of each instance.

Previous methods mainly seek text instances (characters, words or text lines) in local regions, with sliding-window~\cite{Ref:Kim2003, Ref:Chen2004, Ref:Wang2010, Ref:Neumann2013B, Ref:Jaderberg2014} or connected component extraction~\cite{Ref:Epshtein2010,  Ref:Neumann2011, Ref:Yao2012, Ref:Neumann2015} techniques. These algorithms have brought novel ideas into this field and constantly advanced the state-of-the-art. However, most of the existing algorithms spot text within local regions (up to text line level), making it nearly impossible to exploit context in wider scope, which can be critical for dealing with challenging situations. Consequently, they would struggle in hunting weak text instances and suppressing difficult false positives.

Moreover, almost all the previous methods (except for~\cite{Ref:Yao2012, Ref:Kang2014, Ref:Yin2015}) have focused on detecting horizontal or near-horizontal texts, overlooking non-horizontal ones. This largely limits the practicability and adaptability of these methods, since crucial information in regard to the scene might be embodied in such non-horizontal texts.

\begin{figure}[!tp]
\centering
\includegraphics[width=0.95\linewidth]{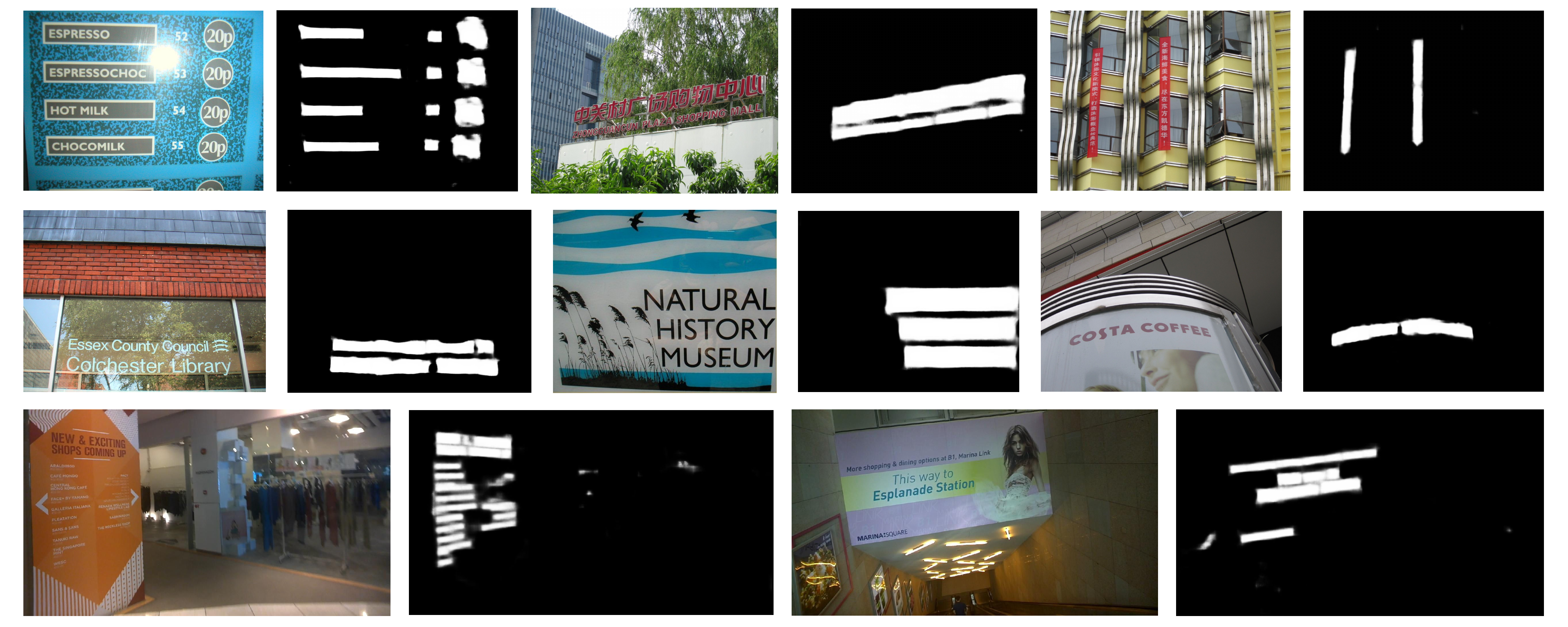}
\caption{Text regions predicted by the proposed text detection algorithm. In this work, scene text detection is casted as a semantic segmentation problem, which is conceptionally and functionally different from previous sliding-window or connected component based approaches.}   \label{Fig:Headline}
\vspace{-4mm}
\end{figure}

\begin{figure}[!tp]
\centering
\includegraphics[width=0.9\linewidth]{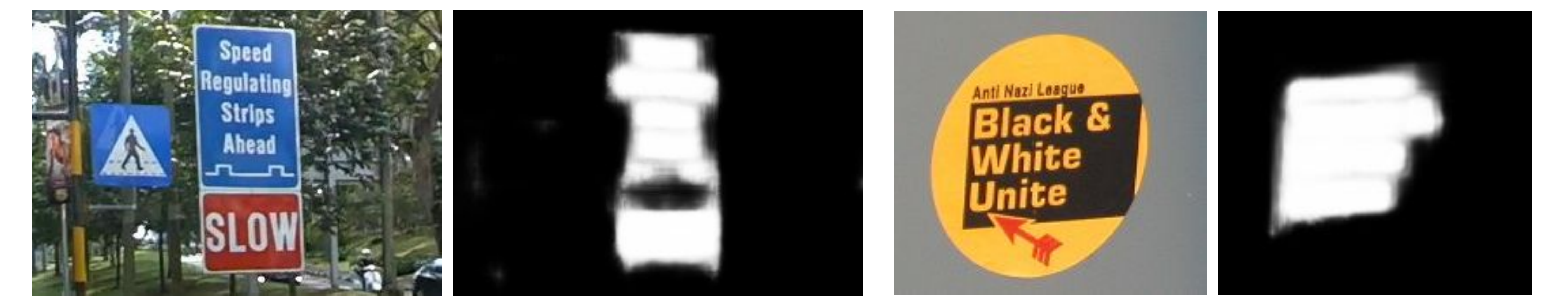}
\caption{An issue caused by simple two-class semantic segmentation. When several text lines are very close, the estimated text regions may stick together, making it difficult to identify each individual text line.}   \label{Fig:Ambiguity}
\vspace{-4mm}
\end{figure}

We propose in this work a novel algorithm for scene text detection, which treats text detection as semantic segmentation problem~\cite{Ref:Shotton2008}. This algorithm performs holistic,  per-pixel estimation and produces dense maps, in which the properties of scene text are implied, as shown in Fig.~\ref{Fig:Headline}. Nevertheless, simple two-class (text \textit{vs.} non-text) semantic segmentation is not sufficient for scene text detection, since multiple text instances that are very close to each other may make it hard to separate each instance (see Fig.~\ref{Fig:Ambiguity}). We tackle this issue by taking into account the center and scale of individual characters as well as the linking orientation of nearby characters, in addition to the location of text regions.

When multiple text lines flock together and their orientations are unknown in advance, it is not trivial to identify and group each text line. Previous methods either simply assume that text lines are horizontal or near-horizontal, or use heuristics to perform text line grouping. In this paper, we propose to build a graph using predicted properties of characters (location, scale, linking orientation, etc) and form text lines with graph partition~\cite{Ref:Buluç2015}.

The proposed strategy is realized using the FCN framework~\cite{Ref:Long2015}, which was originally designed for semantic segmentation. This framework is chosen because it applies multi-scale learning and prediction, conforming to the multi-scale nature of text in natural scenes, and includes rich prior information of natural images, by pretraining on large volume of real-world data (ImageNet~\cite{Ref:Deng2009}). Building upon FCN, the proposed method is able to effectively detect text instances with high variability while coping with hard false alarms in real-world scenarios.

To evaluate the effectiveness and advantages of the proposed algorithm, we have conducted experiments on public datasets in this area, including ICDAR 2013~\cite{Ref:Karatzas2013}, ICDAR 2015~\cite{Ref:Karatzas2015} and MSRA-TD500~\cite{Ref:Yao2012}. The quantitative results demonstrate that the proposed method significantly outperforms previous state-of-the-art algorithms. Specifically, this is the first work\footnote{The algorithms evaluated in~\cite{Ref:Veit2016} were used in the process of data annotation, thus they cannot be considered as valid baselines. We learn about this via correspondence with the authors of~\cite{Ref:Veit2016}.} that reports quantitative performance on the large-scale benchmark COCO-Text~\cite{Ref:Veit2016}, which is much larger (63,686 natural images) and exhibits far more variability and complexity than previous datasets.

In summary, the contributions of this paper are four-folds:

(1) We cast scene text detection as a semantic segmentation problem and make holistic prediction in the detection procedure, in contrast to previous approaches which mainly make decision locally and thus cannot make full use of the contextual information in the whole image.

(2) This work simultaneously predicts the probability of text regions, characters and the relationship among adjacent characters in a unified framework, excavating more properties of scene text and endowing the system with the ability to detect multi-oriented and curved text.

(3) The algorithm substantially outperforms the prior arts on standard benchmarks in this field.

The rest of the paper is structured as follows: We review previous ideas and approaches in Sec.~\ref{Sec:RelatedWork}. The main idea and details of the proposed algorithm are explained in Sec~\ref{Sec:Methodology} and the experiments and comparisons are presented Sec.~\ref{Sec:Experiments}. Conclusion remarks are given in Sec~\ref{Sec:Conclusions}.

\section{Related Work}  \label{Sec:RelatedWork}

Scene text detection and recognition have been extensively studied for a few years in the computer vision community and document analysis community, and plenty of excellent works and effective strategies have been proposed~\cite{Ref:Chen2004, Ref:Epshtein2010, Ref:Neumann2010, Ref:Pan2011, Ref:Mishra2012, Ref:Novikova2012, Ref:Weinman2013, Ref:Bissacco2013, Ref:Jaderberg2014, Ref:Yin2015}. Comprehensive and detailed reviews can be found in~\cite{Ref:Chen2000, Ref:Jung2004, Ref:Uchida2014, Ref:Ye2014}. In this section, we will concentrate on text detection approaches that are most related to the proposed method.

Stroke Width Transform (SWT)~\cite{Ref:Epshtein2010}, Maximally Stable Extremal Regions (MSER)~\cite{Ref:Neumann2010} as well as their variants~\cite{Ref:Yao2012, Ref:Huang2013, Ref:Neumann2012, Ref:Yin2014} have been the mainstream in scene text detection. These methods generally hunt character candidates via edge detection or extreme region extraction. Different from such component-based approaches, Neumann and Matas~\cite{Ref:Neumann2013B} proposed to seek character strokes in a multi-scale sliding-window manner. Zhang~\emph{et al.}~\cite{Ref:Zhang2015} presented a text detector that makes use of the local symmetry property of character groups. The work of Jaderberg~\emph{et al.}~\cite{Ref:Jaderberg2015B} adopted object proposal and regression techniques to spot words in natural images, drawing inspiration from R-CNN~\cite{Ref:Girshick2014}. However, a common issue with these methods is that they all only use cues from local regions (up to text line level) in text detection. In contrast, the proposed algorithm makes decision in a much wider scope (up to whole image), and thus is able to take advantage of both short-range and long-range information in the images, which can be very useful in suppressing false alarms in complex scenes.

Most of the previous methods have focused on horizontal or near-horizontal text, with very few exceptions~\cite{Ref:Yao2012, Ref:Kang2014, Ref:Yin2015}. However, text in real-world situations can be in any orientation. The ignorance of non-horizontal text can be a severe drawback, since important information regarding the scene may be embodied in non-horizontal text. The proposed algorithm, directly inferring the orientation property of each text instance, can naturally and effortlessly deal with text of arbitrary directions.

In recent years, deep learning based methods for scene text detection~\cite{Ref:Coates2011, Ref:Jaderberg2014, Ref:Huang2014, Ref:Jaderberg2015B} have been very popular, due to their advantage in performance over conventional strategies~\cite{Ref:Chen2004, Ref:Epshtein2010, Ref:Neumann2010, Ref:Yi2011, Ref:Huang2013}. The approach proposed in this paper also utilizes deep convolutional neural networks to detect text with high variability and to eliminate false positives caused by complex background. The main difference lies in that the proposed approach works in a holistic fashion and produces global, pixel-wise prediction maps, while the other deep learning based methods essentially perform classification on local regions (image patches~\cite{Ref:Coates2011, Ref:Jaderberg2014} or proposals generated by other techniques~\cite{Ref:Huang2014, Ref:Jaderberg2015B}).

This work is mainly inspired by the Holistically-Nested Edge Detection (HED) method proposed by Xie and Tu~\cite{Ref:Xie2015}. HED is an edge detection algorithm that adopts multi-scale and multi-level feature learning and performs holistic prediction. HED leverages the power of both Fully Convolutional Networks (FCN)~\cite{Ref:Long2015} and Deeply-Supervised Nets (DSN), and obtains significantly enhanced performance on edge detection in natural images. There are mainly three reasons that we adopt HED as the base model of our text detection algorithm: (1) Text is highly correlated with edge, as pointed out in previous works~\cite{Ref:Epshtein2010, Ref:Lee2010, Ref:Chen2011}. (2) HED makes holistic, image-to-image prediction, which allows for the use of wide-scope and long-range contextual information to effectively suppress false positives. (3) HED can directly handle edges of different scales and orientations, which fit well the multi-scale and multi-orientation nature of text in scene images. 

\section{Methodology}  \label{Sec:Methodology}

In this section, we will present the main idea, network architecture and details of the proposed method. Specifically, the pipeline for scene text detection is described in Sec.~\ref{Sec:Pipeline}, after an overview of the whole system (Sec~\ref{Sec:Overview}). The network architecture as well as the notation and formulation of the system are given in Sec.~\ref{Sec:Architecture} and Sec.~\ref{Sec:Formulation}, respectively.
s
\subsection{Overview}    \label{Sec:Overview}

Generally, the proposed algorithm follows the paradigm of the FCN~\cite{Ref:Long2015} and HED framework~\cite{Ref:Xie2015}, i.e., it infers properties of scene text in a holistic fashion by producing image-level, pixel-wise prediction maps. In this paper, we consider text regions (words or text lines), individual characters and their relationships (linking orientation between characters) as the three properties to be estimated at runtime, since these properties are effective for scene text detection and. Moreover, the ground truth of these properties can be easily obtained from standard benchmark datasets. After the pixel-wise prediction maps are generated, detections of scene text are formed by aggregating cues in these maps. As this idea is general, other useful properties regarding scene text (for instance, binary mask of character strokes and script type of text) can be readily introduced into this framework, which are expected to further improve the accuracy of scene text detection.


\subsection{Text Detection Pipeline}  \label{Sec:Pipeline}

\begin{figure*}[!tp]
\centering
\includegraphics[width=1.0\linewidth]{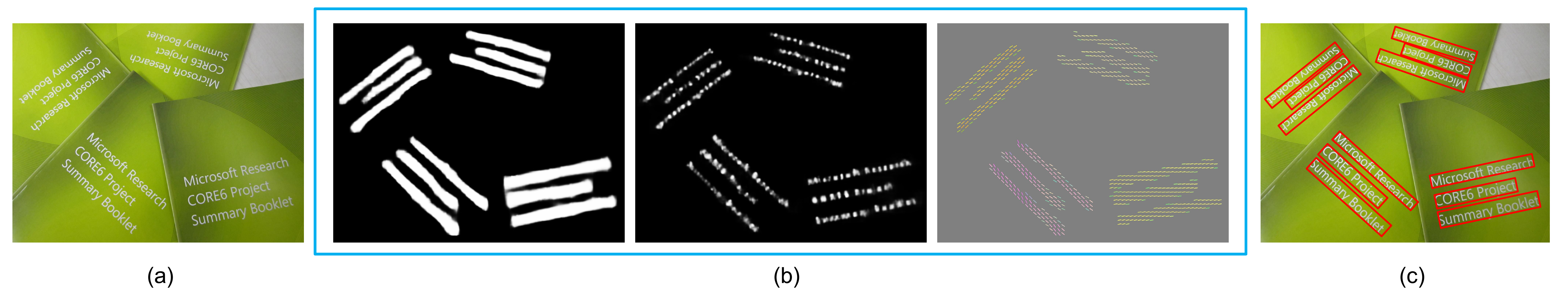}
\caption{Pipeline of the proposed algorithm. (a) Original image. (b) Prediction maps. From left to right: text region map, character map and linking orientation map. For better visualization, linking orientations are represented with color-coded lines and only those within text regions are shown. (c) Detections.}   \label{Fig:Pipeline}
\vspace{-4mm}
\end{figure*}

Unlike previous methods for scene text detection, which usually start from extracting connected components~\cite{Ref:Epshtein2010, Ref:Neumann2010, Ref:Huang2013} or scanning multi-scale windows~\cite{Ref:Chen2004, Ref:Wang2011, Ref:Jaderberg2014} within images, the proposed algorithm operates in an alternative way. As shown in Fig.~\ref{Fig:Pipeline}, the pipeline is quite straightforward: The original image is fed into the trained model and three prediction maps, corresponding to text regions, characters and linking orientations of adjacent characters, are produced. Detections are formed by performing segmentation, aggregation and partition on the three maps. If required, word partition, the process of splitting text lines into individual words, is applied to the previously formed detections.

\subsubsection{Ground Truth Preparation}

\begin{figure}[!tp]
\centering
\includegraphics[width=0.9\linewidth]{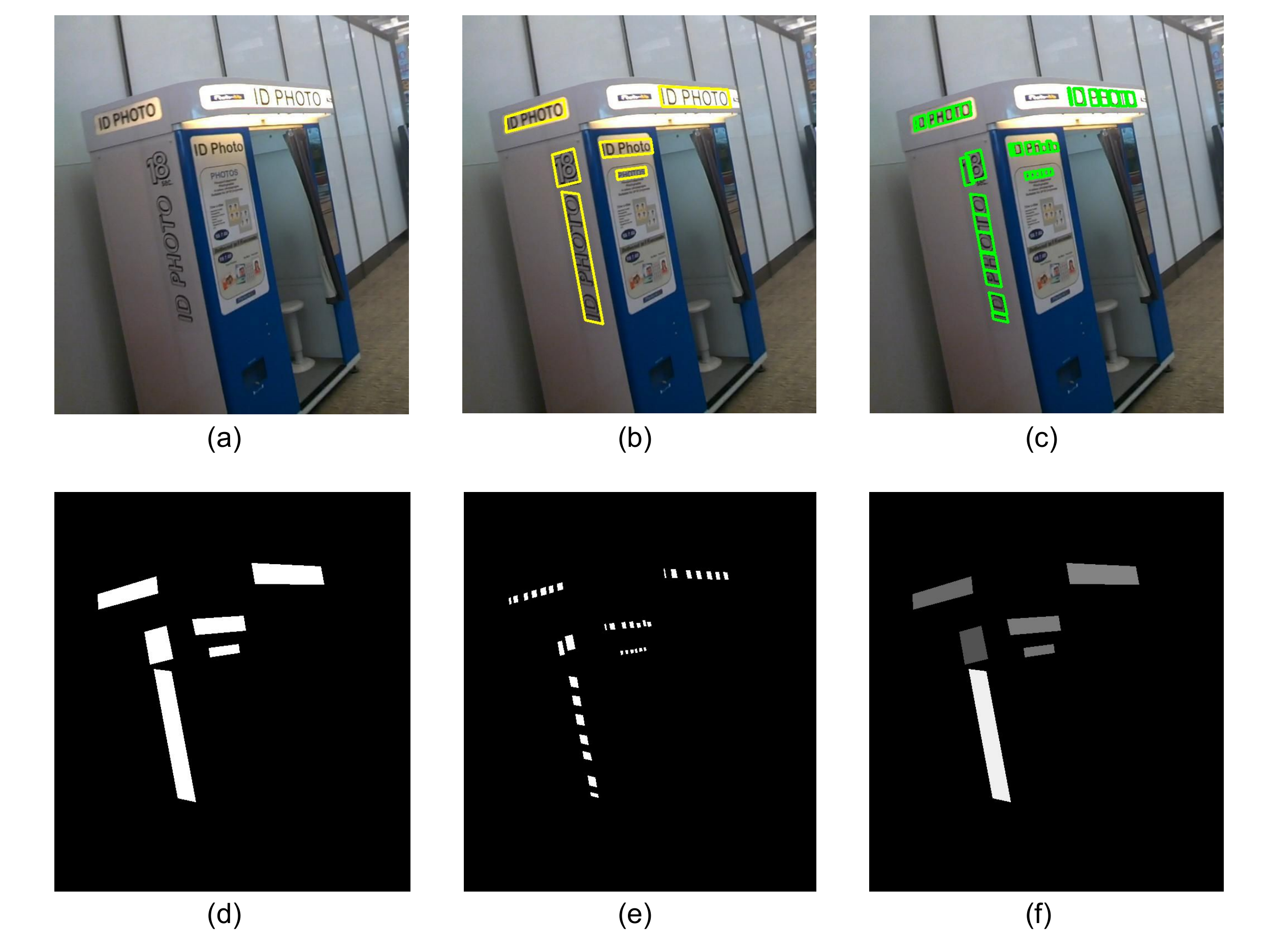}
\vspace{-4mm}
\caption{Ground truth preparation. (a) Original image. (b) Ground truth polygons of text regions. (c) Ground truth polygons of characters. (d) Ground truth map for text regions. (e) Ground truth map for characters. (f) Ground truth map for linking orientations.}  \label{Fig:GT}
\vspace{-4mm}
\end{figure}

Since the problem of scene text detection has been converted to semantic segmentation and the base framework is FCN/HED, the ground truth should be converted to label maps that are compatible with FCN/HED. As depicted in Fig.~\ref{Fig:GT}, for each image three label maps are produced from the ground truth annotations. The label map for text regions is a binary map, in which foreground pixels (i.e., those within text regions) are marked as '1' and background pixels '0'. The label map for characters is a binary map, in which foreground pixels (i.e., those within character regions) are marked as '1' and background pixels '0'. To avoid the situation that the characters in the prediction map stuck together at runtime, the binary masks are shrunk to half of its original size in both dimensions.

The label map for linking orientations is a soft map, in which each foreground pixel is assigned with to a value within the range of $[0,1]$. The orientations of the foreground pixels is assigned as the orientation of the corresponding ground truth polygons of text regions. In this work, we consider linking orientation $\theta$ in the range of $[-\pi/2, \pi/2]$. Linking orientations beyond this range is converted to it. All linking orientations are mapped to $[0,1]$ by shifting and normalization.

\subsubsection{Model Training}

The prediction model is trained by feeding the training images and the corresponding ground truth maps into the network, which will be described in detail in Sec.~\ref{Sec:Architecture}. The training procedure generally follows that of the HED method~\cite{Ref:Xie2015} and the differences will be explained in Sec.~\ref{Sec:Training}. 

\subsubsection{Prediction Map Generation}

With the trained model, maps that represent the information of text regions, characters and linking orientations are generated by inputting the original image into it. For details, see Sec.~\ref{Sec:Testing}. Exemplar prediction maps generated by the trained model are demonstrated in Fig.~\ref{Fig:Pipeline} (b) and Fig.~\ref{Fig:Formation} (b), (c) and (d). Note that only the linking orientations within foreground regions are valid, so those in background regions (probability of being text region less than 0.5) are not shown.

\subsubsection{Detection Formation}

\begin{figure}[!tp]
\centering
\includegraphics[width=0.85\linewidth]{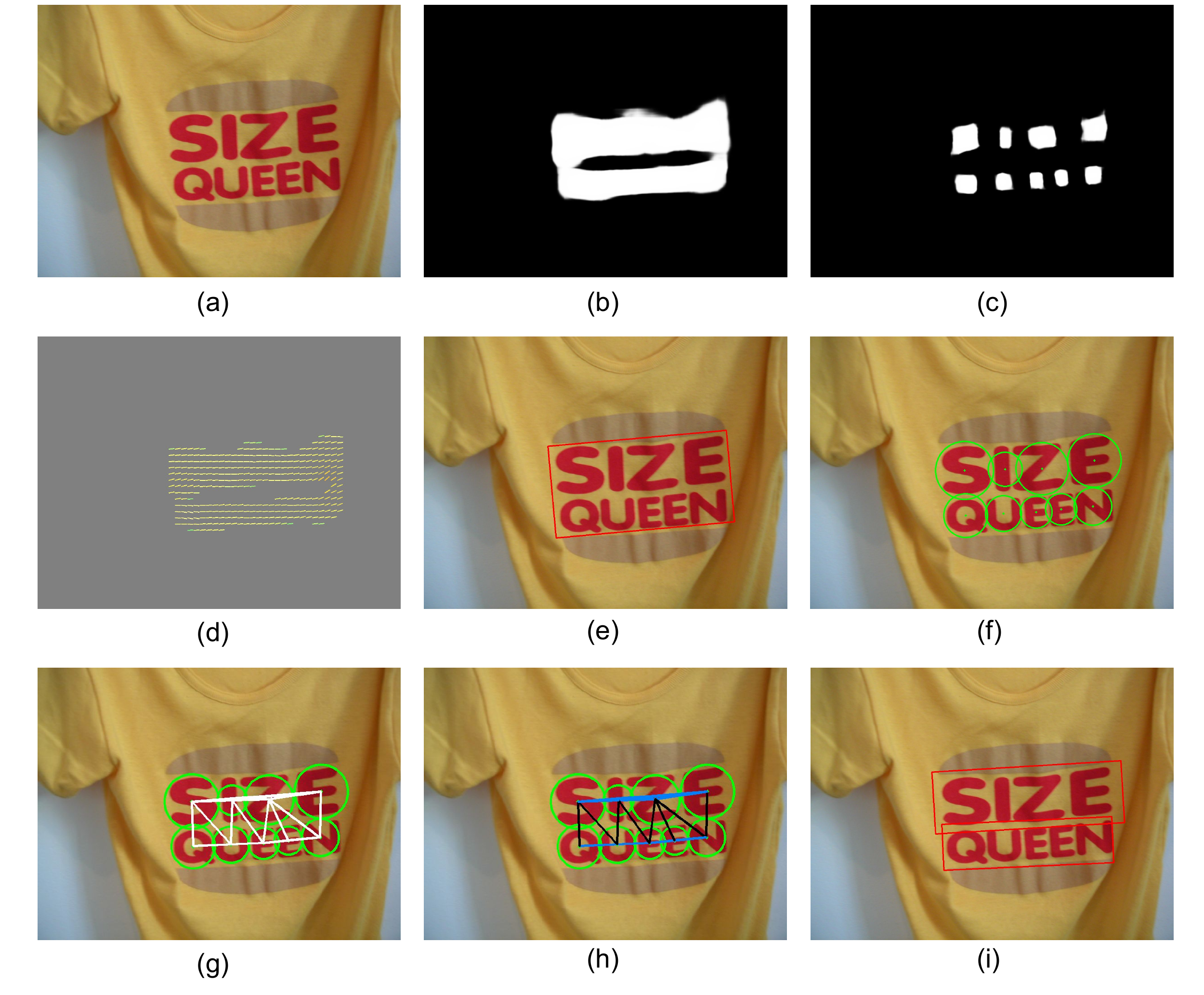}
\vspace{-4mm}
\caption{Detection formation. (a) Original image. (b) Prediction map of text regions. (c) Prediction map of characters. (d) Prediction map of linking orientations. (e) Text region (red rectangle). (f) Characters (green circles). The center and radius of the circles represent the location and scale of the corresponding characters. (g) Delaunay triangulation. (h) Graph partition. Blue lines: linkings retained. Black lines: linkings eliminated. (i) Detections.}  \label{Fig:Formation}
\vspace{-4mm}
\end{figure}

Candidates of text regions and characters are hunted by segmenting the corresponding prediction map with adaptive thresholding (see Fig.~\ref{Fig:Formation} (e) and (f)). Note that since the scales of characters are shrunk to half of their original scales in the training phase, the estimated radii of the character candidates are multiplied with a factor of 2.

Then, character candidates within the same text regions are grouped into cliques and represented as graphs. Given a clique of characters $U={u_{i}, i=1,\cdots,m}$, where $m$ is the character count, Delaunay triangulation~\cite{Ref:Kang2014} is applied to these characters. Delaunay triangulation provides an effective way to eliminate unnecessary linkings of distant characters. The triangulation $T$ is used to construct a graph $G=(U,E)$, in which the vertexes are the characters $U$ and the edges $E$ model the similarities between pairs of characters.

The weight $w$ of edge $e=(u_{i}, u_{j})$ is defined as:
\begin{equation}
w=
\begin{cases}
s(i,j), & \text{if}\ e \in T \\
0, & \text{otherwise}
\end{cases},  \label{Eqn:Weight}
\end{equation}
where $s(i,j)$ is the similarity between the pair of characters $u_{i}$ and $u_{j}$. According to Eqn.~\ref{Eqn:Weight}, the weights of the linkings that not belonging to the triangulation are set to zero.

The similarity $s(i,j)$ between $u_{i}$ and $u_{j}$ is the harmonic average of the spatial similarity and orientation similarity:
\begin{equation}
s(i,j) = \dfrac{2a(i,j)o(i,j)}{a(i,j)+o(i,j)},
\end{equation}
where $a(i,j)$ and $o(i,j)$ denote the spatial similarity and orientation similarity between $u_{i}$ and $u_{j}$.

The spatial similarity $a(i,j)$ defined as:
\begin{equation}
a(i,j) = \exp(-\dfrac{d^2(i,j)}{2D^2}),
\end{equation}
where $d(i,j)$ is the distance of the centers of $u_{i}$ and $u_{j}$, while $D$ is the average length of all the edges in the Delaunay triangulation $T$.

The orientation similarity $o(i,j)$ defined as:
\begin{equation}
o(i,j) = cos(\Lambda(\phi(i,j)-\psi(i,j))),
\end{equation}
where $\phi(i,j)$ is the orientation of the line between the centers of $u_{i}$ and $u_{j}$, while $\psi(i,j)$ is the average value in the area between $u_{i}$ and $u_{j}$ in the linking orientation map (see~\ref{Fig:Formation} (d)). $\Lambda$ is an operator for computing the included angle (acute angle) of two orientations. This definition of orientation similarity rewards pairs that are in accordance with the predicted linking orientations while punishes those that violate such prediction.

To split groups of character candidates into text lines, a simple yet effective strategy proposed by Yin~\emph{et al.}\cite{Ref:Yin2007} is adopted. Based on the graph $G=(U,E)$, a Maximum Spanning Tree~\cite{Ref:Pemmaraju2003}, $M$, is constructed. Selecting and eliminating edges in $M$ is actually performing graph partition on $G$, leading to segmentation of text lines. For example, eliminating any edge in $M$ will partition it into two parts, corresponding two text lines, while eliminating two edges will result in three text lines.

Since the number of text lines is unknown, it should be inferred in the procedure of text line segmentation. Under the assumption that text lines in real-world scenarios are in linear or near-linear form, a straightness measure function is defined:
\begin{equation}
S_{vm}=\sum_{i=1}^{K}\dfrac{\lambda_{i1}}{\lambda_{i2}},
\end{equation}
where $K$ is the number of clusters (text lines), $\lambda_{i1}$ and $\lambda_{i2}$ are the largest and second largest eigenvalues of the covariance matrix $C_{i}$. $C_{i}$ is computed using the coordinates of the centers of characters the in the $i$th cluster. The optimal segmentation of text lines is achieved, when the value of function $S_{vm}$ reaches its maximum.

However, this solution is not applicable to text lines with curved shapes (for example, those in Fig.~\ref{Fig:Curved}), which they violate the linearity assumption. To tackle this problem, a threshold $\tau$ is introduced in this work. While performing text line segmentation, the edges with weight greater than $\tau$ will not be selected or eliminated. Since the proposed method is able to predict linking orientations of characters in straight text lines (see Fig.~\ref{Fig:Pipeline}) as well as in curved text lines (see Fig.~\ref{Fig:Curved}), this strategy works well on text lines with linear shapes and curved shapes.

\subsubsection{Post-Processing}

In certain tasks, such as ICDAR 2013~\cite{Ref:Karatzas2013} and ICDAR 2015\cite{Ref:Karatzas2015}, word partition is required, since text in images from these datasets is labelled in word level. We adopted the word partition method in~\cite{Ref:Epshtein2010}, as it has proven to be simple and effective.

\subsection{Architecture}  \label{Sec:Architecture}

\begin{figure*}[!tp]
\centering
\vspace{-3mm}
\includegraphics[width=0.75\linewidth]{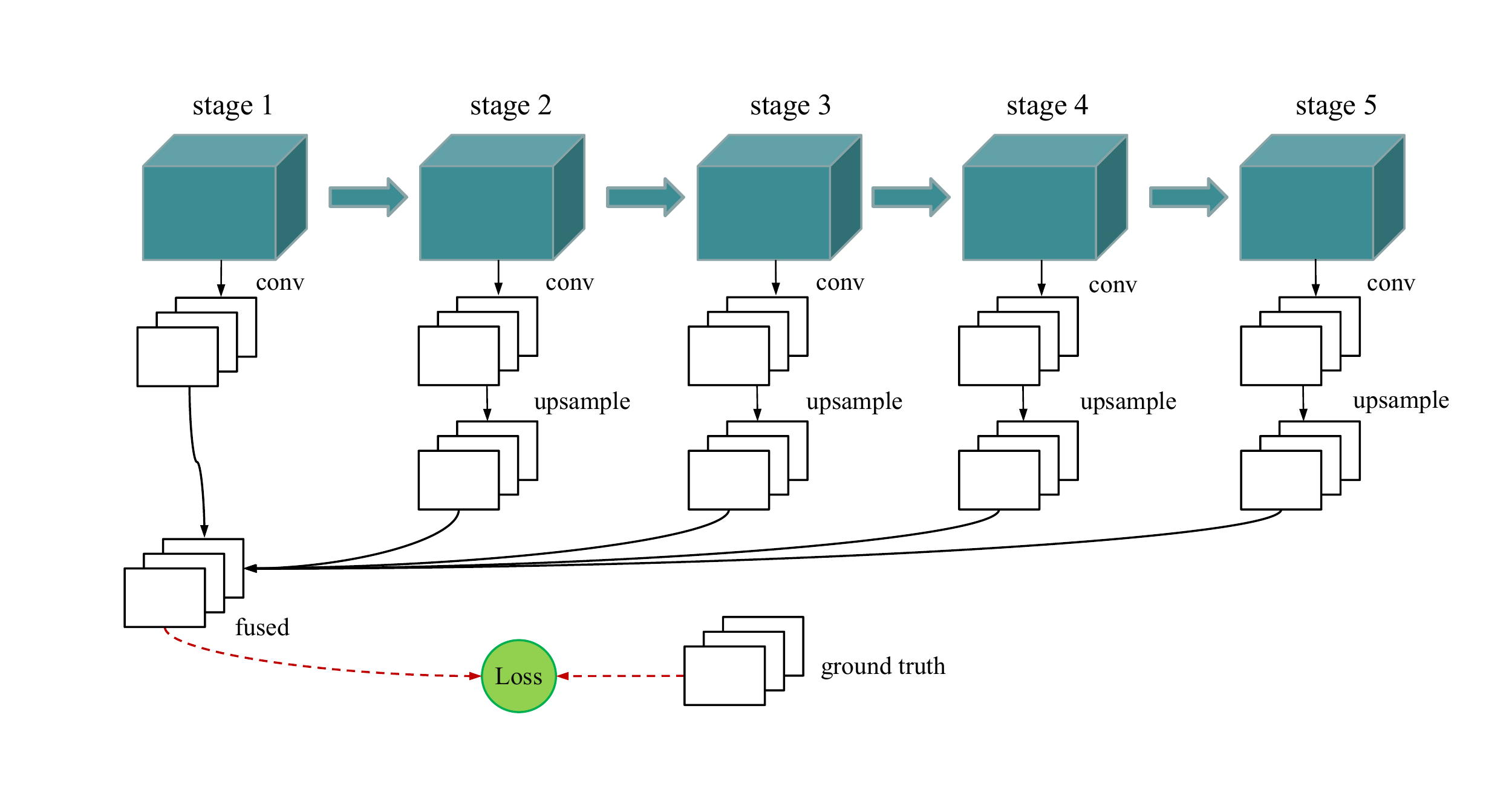}
\vspace{-8mm}
\caption{Network architecture of the proposed algorithm. The base network is inherited from HED~\cite{Ref:Xie2015}, which made surgery on the pretrained VGG-16 Net model~\cite{Ref:Simonyan2015}.}   \label{Fig:Architecture}
\vspace{-5mm}
\end{figure*}

As shown in Fig.~\ref{Fig:Architecture}, the architecture used in this work is based on that of HED~\cite{Ref:Xie2015}, which made surgery on the  pretrained 5-stage VGG-16 Net model~\cite{Ref:Simonyan2015}. Necessary modifications are applied to the architecture of HED, to accomplish the task of text detection. Specifically, three side-output layers, corresponding to text region, character and linking orientation predictions, are connected to the last convolutional layer of each stage. This means that conv1\_2, conv2\_2, conv3\_3, conv4\_3 and conv5\_3 in the network have side-output layers. The side-output layer for text region prediction, character prediction and that for linking orientation prediction are learned independently. Note that from stage 2 to 5, upsampling operation, which is realized by in-network deconvolution, is needed to rescale the prediction maps to the same size of the original image. The outputs of the side layers from the 5 stages are fused to form overall predictions, which is accomplished by another convolutional layer, the same as in~\cite{Ref:Xie2015}. The loss of each stage is computed as described in Sec.~\ref{Sec:Training}.

\subsection{Formulation}   \label{Sec:Formulation}

The notation and formulation of the proposed method follow those of~\cite{Ref:Xie2015}. Here we will keep consistent with~\cite{Ref:Xie2015} and emphasis on the modifications and differences.

\subsubsection{Training Phase}   \label{Sec:Training}

Assume the training set is  $S=\{(X_{n}, Y_{n}),n=1,\cdots,N\}$, where $N$ is the size of the training set, $X_{n}$ is the $n$th original image and $Y_{n}$ is the corresponding ground truth. Different from HED, in which the ground truth $Y_{n}$ is a binary edge map, $Y_{n}$ in this work is composed of three maps, i.e., $Y_{n}=\{R_{n},C_{n},\Theta_{n}\}$, where $R_{n}=\{r_{j}^{(n)}\in\{0,1\}, j=1,\cdots,|R_{n}|\}$ is a binary map that indicates the presence of text regions at each pixel in the original image $X_{n}$, $C_{n}=\{c_{j}^{(n)}\in\{0,1\}, j=1,\cdots,|C_{n}|\}$ is a binary map indicates the presence of characters (shrunk version are used in this work), and $\Theta_{n}=\{\theta_{j}^{(n)}\in[0, 1], j=1,\cdots,|\Theta_{n}|\}$ is a soft map that represents the linking orientations between adjacent characters in text regions. Note that the value of $\theta{j}^{(n)}$ is valid only if $r_{j}^{(n)}=1$, since linking orientation of adjacent characters is undefined in background. For simplicity, the index $n$ will be omitted hereafter, because the model treats each image independently.

The objective function is defined as the loss in the fused outputs, since the losses in lower side-output layers make little difference according to experiments:
\begin{equation}
\mathcal{L}=\mathcal{L}_{\mathrm{fuse}}(\mathbf{W},\mathbf{w}),
\end{equation}
where $\mathcal{L}_{\mathrm{fuse}}(\mathbf{W},\mathbf{w},\mathbf{h})$ is the loss function of the fused outputs. $\mathbf{W}$ and $\mathbf{w}$ are the collection of parameters of all standard layers and those with the side-output layers and fuse layer, respectively. For more details, refer to~\cite{Ref:Xie2015}.

In HED~\cite{Ref:Xie2015}, each stage produces only one prediction map for edges. In this paper, however, each stage is used to generate three prediction maps: one for text regions, one for characters and one for linking orientations. Therefore, each stage is connected to two side-output layers, instead of one. Accordingly, the definition of the loss function is different. $\mathcal{L}_{\mathrm{fuse}}(\mathbf{W},\mathbf{w})$ is a weighted sum of the corresponding channels for the three types of targets:
\begin{multline}
\mathcal{L}_{\mathrm{fuse}}(\mathbf{W},\mathbf{w})=\lambda_{1}\Delta_{r}(\mathbf{W},\mathbf{w})+\\
\lambda_{2}\Delta_{c}(\mathbf{W},\mathbf{w})+\lambda_{3}\Delta_{o}(\mathbf{W},\mathbf{w}),
\end{multline}
where $\Delta_{r}(\mathbf{W},\mathbf{w})$, $\Delta_{c}(\mathbf{W},\mathbf{w})$ and $\Delta_{o}(\mathbf{W},\mathbf{w})$ are the losses in predicting text regions, characters and linking orientations, respectively. $\lambda_{1}$, $\lambda_{2}$ and $\lambda_{3}$ are parameters to control the relative contributions of these three types of loss functions and $\lambda_{1}$+$\lambda_{2}$+$\lambda_{3}=1$.

For an image, assume the ground truth is $Y=\{R,C,\Theta\}$ and the prediction maps produced by the fuse layer are $\hat{R}$, $\hat{C}$ and $\hat{\Theta}$, the loss function for text regions $\Delta_{r}(\mathbf{W},\mathbf{w})$ is similar with Eqn. (2) in~\cite{Ref:Xie2015}:
\begin{multline}
\Delta_{r}(\hat{R},R;\mathbf{W},\mathbf{w})=\\
-\beta\sum_{j=1}^{|R|}R_{j}\log Pr(\hat{R_{j}}=1;\mathbf{W},\mathbf{w})\\
-(1-\beta)\sum_{j=1}^{|R|}(1-R_{j})\log Pr(\hat{R_{j}}=0;\mathbf{W},\mathbf{w}).
\end{multline}
$\beta$ is a class-balancing parameter and it is defined as $\beta=\frac{|R_{-}|}{|R|}$. $|R_{-}|$ denotes the count of pixels in non-text regions and $|R|$ denotes the total count of pixels. The definition of $\Delta_{c}$ is similar with $\Delta_{r}$ and thus is skipped here.

$\Delta_{o}(\mathbf{W},\mathbf{w})$ is defined as:
\begin{equation}
\Delta_{o}(\hat{\Theta},\Theta;R,\mathbf{W},\mathbf{w})=\sum_{j=1}^{|R|}R_{j}(\sin(\pi|\hat{\Theta}_{j}-\Theta_{j}|)).  \label{Eqn:OrientationLoss}
\end{equation}

This loss function has a double-peak shape. When the difference (included angle) between the estimated orientation and true orientation is small, the loss is close to $0$; when it is near $\pi/2$ (or $-\pi/2$), the loss approaches to $1$; when it goes beyond $\pi/2$ (or below $-\pi/2$), the loss deceases since the difference turns around. According to the definition in Eqn.~\ref{Eqn:OrientationLoss}, orientation errors at background pixels will not be taken into account.

\subsubsection{Testing Phase}  \label{Sec:Testing}

In the testing phase, the test image $I$ is mean-subtracted and fed to the trained model. The prediction maps for text regions, characters and linking orientations are obtained by taking the output of the fusion layer:

\begin{equation}
(\hat{R}, \hat{C}, \hat{\Theta})=\hat{Y}_{\mathrm{fuse}}
\end{equation}

Different from HED, which used the average of the outputs of all the output-layers and the fusion layers, in this work we only employ the responses of the fusion layer. In practice, we found that the outputs of the side-output layers are likely to introduce noises and unimportant details into the prediction maps, which may be harmful to the task of scene text detection.

\section{Experiments and Discussions}  \label{Sec:Experiments}

We implemented the proposed algorithm using the code\footnote{\url{https://github.com/s9xie/hed}} released by the authors of HED~\cite{Ref:Xie2015}, which is based on the Caffe framework~\cite{Ref:Jia2014}. The proposed algorithm was evaluated on three standard benchmarks in this field and compared with other scene text detection methods. All the experiments were conducted on a regular server (2.6GHz 8-core CPU, 128G RAM, Tesla K40m GPU and Linux 64-bit OS) and the routine ran on a single GPU in each time.

\subsection{Datasets}

The datasets used in the experiments will be introduced briefly:

\textbf{ICDAR 2013.} The ICDAR 2013 dataset\footnote{\url{http://rrc.cvc.uab.es/?ch=2&com=downloads}} is from the ICDAR 2013 Robust Reading Competition~\cite{Ref:Karatzas2013}. There are 229 natural images for training and 233 natural images for testing. All the text instances in this dataset are in English and are horizontally placed.

\textbf{ICDAR 2015.} The ICDAR 2015 dataset\footnote{\url{http://rrc.cvc.uab.es/?ch=4&com=downloads}} is from the Challenge 4 (Incidental Scene Text challenge) of the ICDAR 2015 Robust Reading Competition~\cite{Ref:Karatzas2015}. The dataset includes 1500 natural images in total, which are acquired using Google Glass. Different from the images from the previous ICDAR competitions~\cite{Ref:Lucas2005, Ref:Shahab2011, Ref:Karatzas2013}, in which the text instances are well positioned and focused, the images from ICDAR 2015 are taken without user's prior preference or intention, so the text instances are usually skewed or blurred.

\textbf{MSRA-TD500.} The MSRA Text Detection 500 Database (MSRA-TD500)\footnote{\url{http://www.iapr-tc11.org/mediawiki/index.php/MSRA_Text_Detection_500_Database_(MSRA-TD500)}} is a benchmark dataset for assessing detection algorithms for text of different orientations, which is originally proposed in~\cite{Ref:Yao2012}. This dataset has 500 high-resolution natural scene images, in which text may be in varying directions and the language types include both Chinese and English. The training set consists of 300 images and the test set contains 200 images. This dataset is challenging due to the variability of text as well as the complexity of backgrounds.

\textbf{COCO-Text.} The COCO-Text\footnote{\url{http://vision.cornell.edu/se3/coco-text/}} is a newly released, large scale dataset for text detection and recognition in natural images. The original images are from the Microsoft COCO~\cite{Ref:Lin2015} dataset. In COCO-Text, 173,589 text instances from 63,686 images are annotated and each instance has 3 fine-grained text attributes. 43,686 images were chosen by the authors as training set, while 20,000 as validation/test set. This database is the largest benchmark in this area to date.

\subsection{Implementation Details}

We directly made surgery on the network architecture of HED~\cite{Ref:Xie2015} to construct our model for scene text detection, and most parameters were inherited from it. In our model, the learning rate is adjusted to 1e-8 to avoid gradient explosion. Following~\cite{Ref:Xie2015}, we also used VGG-16 Net~\cite{Ref:Simonyan2015} pre-trained on the ImageNet dataset~\cite{Ref:Deng2009}. $\lambda_{1}=\lambda_{2}=\lambda_{3}=\frac{1}{3}$ and $\tau=0.8$ are used in all the experiments.

The training data are the union of the training images from the three datasets ICDAR 2013 (229 training images), ICDAR 2015 (1000 training images) and MSRA-TD500 (300 training images). These 1529 training images were first rescaled to have maximum dimension of 960 pixels with aspect ratio kept unchanged. Then we evenly rotated the images to 36 different angles (with 10 degree angle interval). The corresponding ground truth maps were generated as described in Sec.~\ref{Sec:Methodology} and undergone the same augmentations as the original images. We found that the framework is insensitive to image scale, in accordance with~\cite{Ref:Xie2015}. Therefore, we did not resize the images (as well as the ground truth maps) to multiple scales. In testing, text detection is performed at multiple scales and the detections of different scales are fused to form the final results.

\subsection{Experiments and Discussions}

\subsubsection{Qualitative Results}

\begin{figure*}[!tp]
\centering
\includegraphics[width=0.95\linewidth]{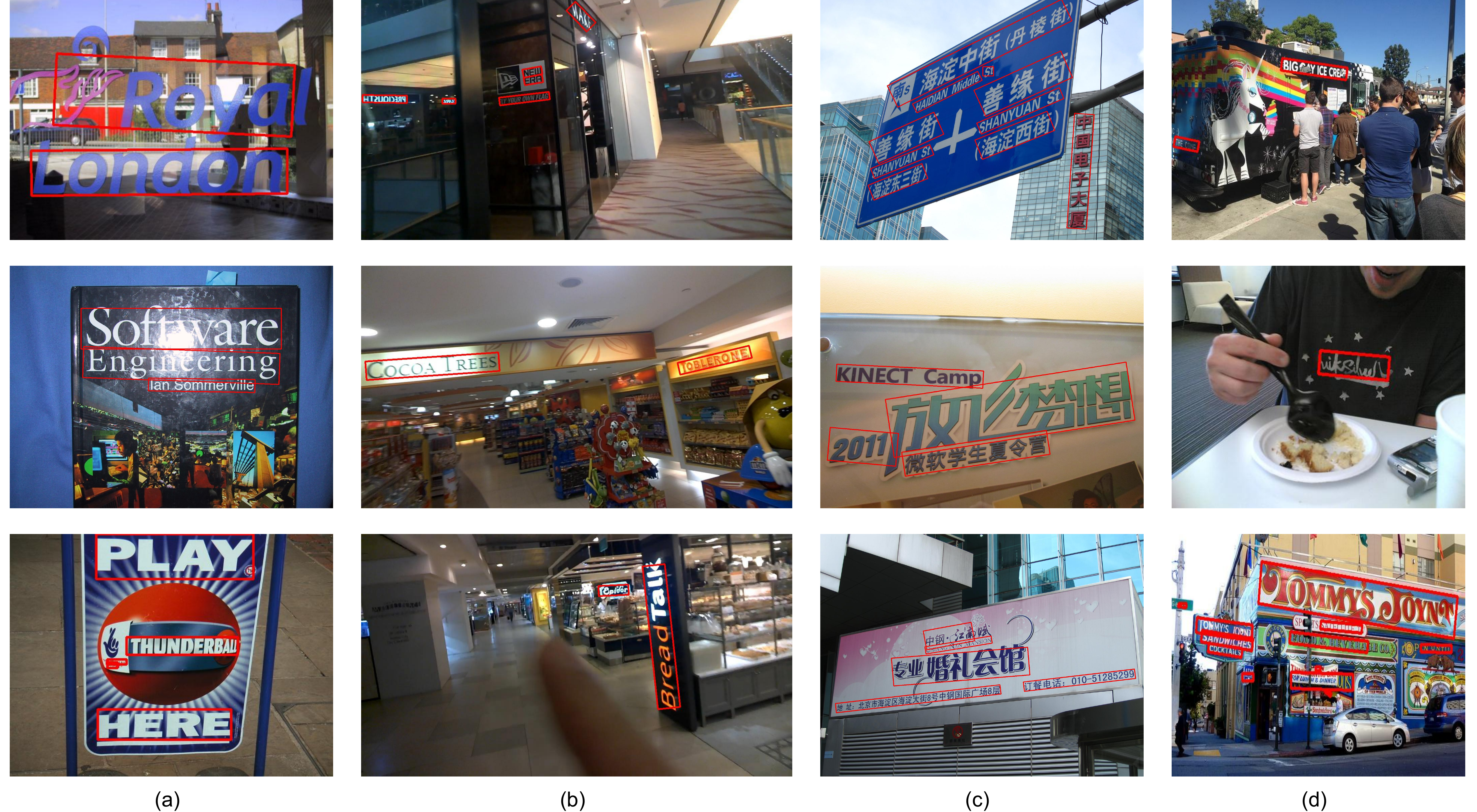}
\caption{Detection examples of the proposed algorithm. (a) ICDAR 2013. (b) ICDAR 2015. (c) MSRA-TD500. (d) COCO-Text}   \label{Fig:Examples}
\vspace{-4mm}
\end{figure*}

Fig.~\ref{Fig:Examples} illustrates a group of detection examples of the proposed algorithm on the four benchmark datasets. As can be seen, the proposed algorithm is able to handle text instances of different orientations, languages, fonts, colors and scales in diverse scenarios. Moreover, it is insensitive to non-uniform illumination, blur, local distractor and connected strokes, to some extent. These examples show the adaptability and robustness of the proposed method.

\subsubsection{Quantitative Results on ICDAR 2013}

\begin{table}
\caption{Performances of different text detection methods evaluated on ICDAR 2013.} 
\label{Tab:ICDAR2013Detection}
\vspace{-3mm}
\begin{center}
\begin{tabular}{|c|c|c|c|}
\hline
\textbf{\begin{footnotesize}Algorithm\end{footnotesize}}&\textbf{\begin{footnotesize}Precision\end{footnotesize}}&\textbf{\begin{footnotesize}Recall\end{footnotesize}}&\textbf{\begin{footnotesize}F-measure\end{footnotesize}}  \\
\hline
\hline
\begin{footnotesize}
Proposed
\end{footnotesize}                  &
\begin{footnotesize}
0.8888
\end{footnotesize} &
\begin{footnotesize}
\textbf{0.8022}
\end{footnotesize} &
\begin{footnotesize}
\textbf{0.8433}
\end{footnotesize} \\
\hline
\begin{footnotesize}
VGGMaxNet~\cite{Ref:Jaderberg2014B}
\end{footnotesize}                  &
\begin{footnotesize}
\textbf{0.9218}
\end{footnotesize} &
\begin{footnotesize}
0.7732
\end{footnotesize} &
\begin{footnotesize}
0.8410
\end{footnotesize} \\
\hline
\begin{footnotesize}
Zhang~\emph{et al.}~\cite{Ref:Zhang2016}
\end{footnotesize}                  &
\begin{footnotesize}
0.88
\end{footnotesize} &
\begin{footnotesize}
0.78
\end{footnotesize} &
\begin{footnotesize}
0.83
\end{footnotesize} \\
\hline
\begin{footnotesize}
Zhang~\emph{et al.}~\cite{Ref:Zhang2015}
\end{footnotesize}                  &
\begin{footnotesize}
0.88
\end{footnotesize} &
\begin{footnotesize}
0.74
\end{footnotesize} &
\begin{footnotesize}
0.80
\end{footnotesize} \\
\hline
\begin{footnotesize}
Tian~\emph{et al.}~\cite{Ref:Tian2015}
\end{footnotesize}                  &
\begin{footnotesize}
0.85
\end{footnotesize} &
\begin{footnotesize}
0.76
\end{footnotesize} &
\begin{footnotesize}
0.80
\end{footnotesize} \\
\hline
\begin{footnotesize}
Lu~\emph{et al.}~\cite{Ref:Lu2015}
\end{footnotesize}                  &
\begin{footnotesize}
0.89
\end{footnotesize} &
\begin{footnotesize}
0.70
\end{footnotesize} &
\begin{footnotesize}
0.78
\end{footnotesize} \\
\hline
\begin{footnotesize}
iwrr2014~\cite{Ref:Zamberletti2015}
\end{footnotesize}                  &
\begin{footnotesize}
0.86
\end{footnotesize} &
\begin{footnotesize}
0.70
\end{footnotesize} &
\begin{footnotesize}
0.77
\end{footnotesize} \\
\hline
\begin{footnotesize}
USTB TexStar~\cite{Ref:Yin2014}
\end{footnotesize}                  &
\begin{footnotesize}
0.88
\end{footnotesize} &
\begin{footnotesize}
0.66
\end{footnotesize} &
\begin{footnotesize}
0.76
\end{footnotesize} \\
\hline
\begin{footnotesize}
Text Spotter~\cite{Ref:Neumann2012}
\end{footnotesize}                  &
\begin{footnotesize}
0.88
\end{footnotesize} &
\begin{footnotesize}
0.65
\end{footnotesize} &
\begin{footnotesize}
0.74
\end{footnotesize} \\
\hline
\begin{footnotesize}
CASIA NLPR~\cite{Ref:Karatzas2013}
\end{footnotesize}                  &
\begin{footnotesize}
0.79
\end{footnotesize} &
\begin{footnotesize}
0.68
\end{footnotesize} &
\begin{footnotesize}
0.73
\end{footnotesize} \\
\hline
\begin{footnotesize}
Text Detector CASIA~\cite{Ref:Shi2013B}
\end{footnotesize}                  &
\begin{footnotesize}
0.85
\end{footnotesize} &
\begin{footnotesize}
0.63
\end{footnotesize} &
\begin{footnotesize}
0.72
\end{footnotesize} \\
\hline
\begin{footnotesize}
I2R NUS FAR~\cite{Ref:Karatzas2013}
\end{footnotesize}                  &
\begin{footnotesize}
0.75
\end{footnotesize} &
\begin{footnotesize}
0.69
\end{footnotesize} &
\begin{footnotesize}
0.72
\end{footnotesize} \\
\hline
\begin{footnotesize}
I2R NUS~\cite{Ref:Karatzas2013}
\end{footnotesize}                  &
\begin{footnotesize}
0.73
\end{footnotesize} &
\begin{footnotesize}
0.66
\end{footnotesize} &
\begin{footnotesize}
0.69
\end{footnotesize} \\
\hline
\begin{footnotesize}
TH-TextLoc~\cite{Ref:Karatzas2013}
\end{footnotesize}                  &
\begin{footnotesize}
0.70
\end{footnotesize} &
\begin{footnotesize}
0.65
\end{footnotesize} &
\begin{footnotesize}
0.67
\end{footnotesize} \\
\hline
\end{tabular}
\end{center}
\vspace{-3mm}
\end{table}

The text detection performance of the proposed algorithm as well as other methods evaluated on the ICDAR 2013 dataset are shown in Tab.~\ref{Tab:ICDAR2013Detection}. The proposed method achieves the highest recall (0.8022) among all the methods. Specifically, the F-measure of the proposed algorithm is slightly better than that of previous state-of-the-art methods~\cite{Ref:Jaderberg2014B}.

Note that on the ICDAR 2013 dataset the performance improvement of our algorithm over previous methods is not as obvious as on the ICDAR 2015 dataset (see Tab.~\ref{Tab:ICDAR2015Detection}) and MSRA-TD500 dataset (see Tab.~\ref{Tab:MSRADetection}). There are two main reasons: (1) The ICDAR 2013 dataset is much more time-honored (most images were from previous ICDAR competitions that can date back to 2003) and most text detection algorithms have saturated on it. (2) The main advantage of our algorithm lies in the capacity of handling multi-oriented text, but vast majority of the text instances in ICDAR 2013 are horizontal. So the advantage of our algorithm cannot be reflected when using ICDAR 2013 as benchmark.

\subsubsection{Quantitative Results on ICDAR 2015}

\begin{table}
\caption{Performances of different text detection methods evaluated on ICDAR 2015.} 
\label{Tab:ICDAR2015Detection}
\vspace{-3mm}
\begin{center}
\begin{tabular}{|c|c|c|c|}
\hline
\textbf{\begin{footnotesize}Algorithm\end{footnotesize}}&\textbf{\begin{footnotesize}Precision\end{footnotesize}}&\textbf{\begin{footnotesize}Recall\end{footnotesize}}&\textbf{\begin{footnotesize}F-measure\end{footnotesize}}  \\
\hline
\hline
\begin{footnotesize}
Proposed
\end{footnotesize}                  &
\begin{footnotesize}
0.7226
\end{footnotesize} &
\begin{footnotesize}
\textbf{0.5869}
\end{footnotesize} &
\begin{footnotesize}
\textbf{0.6477}
\end{footnotesize} \\
\hline
\begin{footnotesize}
Zhang~\emph{et al.}~\cite{Ref:Zhang2016}
\end{footnotesize}                  &
\begin{footnotesize}
0.71
\end{footnotesize} &
\begin{footnotesize}
0.43
\end{footnotesize} &
\begin{footnotesize}
0.54
\end{footnotesize} \\
\hline
\begin{footnotesize}
Stradvision-2~\cite{Ref:Karatzas2015}
\end{footnotesize}                  &
\begin{footnotesize}
\textbf{0.7746}
\end{footnotesize} &
\begin{footnotesize}
0.3674
\end{footnotesize} &
\begin{footnotesize}
0.4984
\end{footnotesize} \\
\hline
\begin{footnotesize}
Stradvision-1~\cite{Ref:Karatzas2015}
\end{footnotesize}                  &
\begin{footnotesize}
0.5339
\end{footnotesize} &
\begin{footnotesize}
0.4627
\end{footnotesize} &
\begin{footnotesize}
0.4957
\end{footnotesize} \\
\hline
\begin{footnotesize}
NJU~\cite{Ref:Karatzas2015}
\end{footnotesize}                  &
\begin{footnotesize}
0.7044
\end{footnotesize} &
\begin{footnotesize}
0.3625
\end{footnotesize} &
\begin{footnotesize}
0.4787
\end{footnotesize} \\
\hline
\begin{footnotesize}
AJOU~\cite{Ref:Koo2013}
\end{footnotesize}                  &
\begin{footnotesize}
0.4726
\end{footnotesize} &
\begin{footnotesize}
0.4694
\end{footnotesize} &
\begin{footnotesize}
0.471
\end{footnotesize} \\
\hline
\begin{footnotesize}
HUST-MCLAB~\cite{Ref:Karatzas2015}
\end{footnotesize}                  &
\begin{footnotesize}
0.44
\end{footnotesize} &
\begin{footnotesize}
0.3779
\end{footnotesize} &
\begin{footnotesize}
0.4066
\end{footnotesize} \\
\hline
\begin{footnotesize}
Deep2Text-MO~\cite{Ref:Yin2014, Ref:Yin2015}
\end{footnotesize}                  &
\begin{footnotesize}
0.4959
\end{footnotesize} &
\begin{footnotesize}
0.3211
\end{footnotesize} &
\begin{footnotesize}
0.3898
\end{footnotesize} \\
\hline
\begin{footnotesize}
CNN MSER~\cite{Ref:Karatzas2015}
\end{footnotesize}                  &
\begin{footnotesize}
0.3471
\end{footnotesize} &
\begin{footnotesize}
0.3442
\end{footnotesize} &
\begin{footnotesize}
0.3457
\end{footnotesize} \\
\hline
\begin{footnotesize}
TextCatcher-2~\cite{Ref:Karatzas2015}
\end{footnotesize}                  &
\begin{footnotesize}
0.2491
\end{footnotesize} &
\begin{footnotesize}
0.3481
\end{footnotesize} &
\begin{footnotesize}
0.2904
\end{footnotesize} \\
\hline
\end{tabular}
\end{center}
\vspace{3mm}
\end{table}

The text detection performance of the proposed method as well as other competing methods on the ICDAR 2015 dataset are shown in Tab.~\ref{Tab:ICDAR2015Detection}. The proposed method achieves the highest recall (0.5869) and the second highest precision (0.7226). Specifically, the F-measure of the proposed algorithm is significantly better than that of previous state-of-the-art~\cite{Ref:Zhang2016} (0.6477 vs. 0.54). This confirms the effectiveness and advantage of the proposed approach.

\subsubsection{Quantitative Results on MSRA-TD500}

\begin{table}
\caption{Performances of different text detection methods evaluated on MSRA-TD500.} 
\label{Tab:MSRADetection}
\vspace{-3mm}
\begin{center}
\begin{tabular}{|c|c|c|c|}
\hline
\textbf{\begin{footnotesize}Algorithm\end{footnotesize}}&\textbf{\begin{footnotesize}Precision\end{footnotesize}}&\textbf{\begin{footnotesize}Recall\end{footnotesize}}&\textbf{\begin{footnotesize}F-measure\end{footnotesize}}  \\
\hline
\hline
\begin{footnotesize}
Proposed
\end{footnotesize}                  &
\begin{footnotesize}
0.7651
\end{footnotesize} &
\begin{footnotesize}
\textbf{0.7531}
\end{footnotesize} &
\begin{footnotesize}
\textbf{0.7591}
\end{footnotesize} \\
\hline
\begin{footnotesize}
Zhang~\emph{et al.}~\cite{Ref:Zhang2016}
\end{footnotesize}                  &
\begin{footnotesize}
\textbf{0.83}
\end{footnotesize} &
\begin{footnotesize}
0.67
\end{footnotesize} &
\begin{footnotesize}
0.74
\end{footnotesize} \\
\hline
\begin{footnotesize}
Yin~\emph{et al.}~\cite{Ref:Yin2015}
\end{footnotesize}                  &
\begin{footnotesize}
0.81
\end{footnotesize} &
\begin{footnotesize}
0.63
\end{footnotesize} &
\begin{footnotesize}
0.71
\end{footnotesize} \\
\hline
\begin{footnotesize}
Kang~\emph{et al.}~\cite{Ref:Kang2014}
\end{footnotesize}                         &
\begin{footnotesize}
0.71
\end{footnotesize}  &
\begin{footnotesize}
0.62
\end{footnotesize} &
\begin{footnotesize}
0.66
\end{footnotesize} \\
\hline
\begin{footnotesize}
Yin~\emph{et al.}~\cite{Ref:Yin2014}
\end{footnotesize}                  &
\begin{footnotesize}
0.71
\end{footnotesize} &
\begin{footnotesize}
0.61
\end{footnotesize} &
\begin{footnotesize}
0.66
\end{footnotesize} \\
\hline
\begin{footnotesize}
Unified~\cite{Ref:Yao2014C}
\end{footnotesize}                         &
\begin{footnotesize}
0.64
\end{footnotesize}  &
\begin{footnotesize}
0.62
\end{footnotesize} &
\begin{footnotesize}
0.61
\end{footnotesize} \\
\hline
\begin{footnotesize}
TD-Mixture~\cite{Ref:Yao2012}
\end{footnotesize}                         &
\begin{footnotesize}
0.63
\end{footnotesize}  &
\begin{footnotesize}
0.63
\end{footnotesize} &
\begin{footnotesize}
0.60
\end{footnotesize} \\
\hline
\begin{footnotesize}
TD-ICDAR~\cite{Ref:Yao2012}
\end{footnotesize}                         &
\begin{footnotesize}
0.53
\end{footnotesize} &
\begin{footnotesize}
0.52
\end{footnotesize} &
\begin{footnotesize}
0.50
\end{footnotesize} \\
\hline
\begin{footnotesize}
Epshtein~\emph{et al.}~\cite{Ref:Epshtein2010}
\end{footnotesize} &
\begin{footnotesize}
0.25
\end{footnotesize}  &
\begin{footnotesize}
0.25
\end{footnotesize} &
\begin{footnotesize}
0.25
\end{footnotesize} \\
\hline
\begin{footnotesize}
Chen~\emph{et al.}~\cite{Ref:Chen2004}
\end{footnotesize} &
\begin{footnotesize}
0.05
\end{footnotesize} &
\begin{footnotesize}
0.05
\end{footnotesize} &
\begin{footnotesize}
0.05
\end{footnotesize} \\
\hline
\end{tabular}
\end{center}
\vspace{-3mm}
\end{table}

The performances of different text detection methods on the MSRA-TD500 dataset~\cite{Ref:Yao2012} are depicted in Tab.~\ref{Tab:MSRADetection}. As can be seen, the proposed algorithm achieves the highest recall and F-measure and the second highest precision. Specially, the F-measure of the proposed algorithm significantly outperforms that of the prior art~\cite{Ref:Zhang2016} (0.7591 vs. 0.74). The improvement in recall is even more obvious (0.7531 vs. 0.63).

\subsubsection{Quantitative Results on COCO-Text}

\begin{table}
\caption{Performances of different text detection methods evaluated on COCO-Text.} 
\label{Tab:COCODetection}
\vspace{-3mm}
\begin{center}
\begin{tabular}{|c|c|c|c|}
\hline
\textbf{\begin{footnotesize}Algorithm\end{footnotesize}}&\textbf{\begin{footnotesize}Precision\end{footnotesize}}&\textbf{\begin{footnotesize}Recall\end{footnotesize}}&\textbf{\begin{footnotesize}F-measure\end{footnotesize}}  \\
\hline
\hline
\begin{footnotesize}
Proposed
\end{footnotesize}    &
\begin{footnotesize}
0.4323
\end{footnotesize} &
\begin{footnotesize}
0.271
\end{footnotesize} &
\begin{footnotesize}
0.3331
\end{footnotesize} \\
\hline
\hline
\multicolumn{4}{|c|}{\textbf{\begin{footnotesize}Baselines from~\cite{Ref:Veit2016}\end{footnotesize}}} \\
\hline
\begin{footnotesize}
A
\end{footnotesize}    &
\begin{footnotesize}
0.8378
\end{footnotesize} &
\begin{footnotesize}
0.233
\end{footnotesize} &
\begin{footnotesize}
0.3648
\end{footnotesize} \\
\hline
\begin{footnotesize}
B
\end{footnotesize}    &
\begin{footnotesize}
0.8973
\end{footnotesize} &
\begin{footnotesize}
0.107
\end{footnotesize} &
\begin{footnotesize}
0.1914
\end{footnotesize} \\
\hline
\begin{footnotesize}
C
\end{footnotesize}    &
\begin{footnotesize}
0.1856
\end{footnotesize} &
\begin{footnotesize}
0.047
\end{footnotesize} &
\begin{footnotesize}
0.0747
\end{footnotesize} \\
\hline
\end{tabular}
\end{center}
\vspace{-3mm}
\end{table}

The performance of the proposed method on the COCO-Text dataset~\cite{Ref:Veit2016} is depicted in Tab.~\ref{Tab:COCODetection}\footnote{The numbers in this table have been modified, since the evaluation script for COCO-Text (\url{https://github.com/andreasveit/coco-text}) has been updated.}. On this large, challenging benchmark, the proposed algorithm achieves 0.4323, 0.271 and 0.3331 in precision, recall and F-measure, respectively. As reference, the performances of the baseline methods from~\cite{Ref:Veit2016} are included. However, it is not possible to directly compare our method with these baselines, since they were used for labelling the ground truth in the data annotation procedure and they were evaluated on the full dataset (63686 image) instead of the validation set (20,000 images).

\begin{table}
\caption{Performances on sub-categories in COCO-Text.} 
\label{Tab:COCOLegible}
\begin{center}
\begin{tabular}{|c|c|c|c|c|}
\hline
\textbf{\begin{footnotesize}Algorithm\end{footnotesize}}&\multicolumn{4}{|c|}{\textbf{\begin{footnotesize}Recall\end{footnotesize}}}  \\
\hline
& \multicolumn{2}{|c|}{\textbf{\begin{footnotesize}Legible\end{footnotesize}}} &\multicolumn{2}{|c|}{\textbf{\begin{footnotesize}Illegible\end{footnotesize}}}  \\
\hline
& \textbf{\begin{footnotesize}Machine\end{footnotesize}} & \textbf{\begin{footnotesize}Hand\end{footnotesize}} & \textbf{\begin{footnotesize}Machine\end{footnotesize}} & \textbf{\begin{footnotesize}Hand\end{footnotesize}} \\
\hline
\hline
\begin{footnotesize}
Proposed
\end{footnotesize}    &
\begin{footnotesize}
0.3563
\end{footnotesize} &
\begin{footnotesize}
0.3109
\end{footnotesize} &
\begin{footnotesize}
0.0297
\end{footnotesize} &
\begin{footnotesize}
0.0567
\end{footnotesize} \\
\hline
\hline
\multicolumn{5}{|c|}{\textbf{\begin{footnotesize}Baselines from~\cite{Ref:Veit2016}\end{footnotesize}}} \\
\hline
\begin{footnotesize}
A
\end{footnotesize}    &
\begin{footnotesize}
0.3401
\end{footnotesize} &
\begin{footnotesize}
0.1511
\end{footnotesize} &
\begin{footnotesize}
0.0409
\end{footnotesize} &
\begin{footnotesize}
0.0231
\end{footnotesize} \\
\hline
\begin{footnotesize}
B
\end{footnotesize}    &
\begin{footnotesize}
0.1616
\end{footnotesize} &
\begin{footnotesize}
0.1096
\end{footnotesize} &
\begin{footnotesize}
0.0069
\end{footnotesize} &
\begin{footnotesize}
0.0022
\end{footnotesize} \\
\hline
\begin{footnotesize}
C
\end{footnotesize}    &
\begin{footnotesize}
0.0709
\end{footnotesize} &
\begin{footnotesize}
0.0463
\end{footnotesize} &
\begin{footnotesize}
0.0028
\end{footnotesize} &
\begin{footnotesize}
0.0022
\end{footnotesize} \\
\hline
\end{tabular}
\end{center}
\vspace{-5mm}
\end{table}

The performance of the proposed method on different sub-categories in the COCO-Text dataset is depicted in Tab.~\ref{Tab:COCOLegible}\footnote{The numbers in this table have been modified, since the evaluation script for COCO-Text (\url{https://github.com/andreasveit/coco-text}) has been updated.}. The proposed algorithm performs much better on legible, machine printed text than illegible, handwritten text. Likewise, the performances of the baseline methods from~\cite{Ref:Veit2016} are also included.

\subsection{Running time of Proposed Method}

On 640x480 images, it take the proposed method about 0.42s on average to produce the prediction maps, when running on a K40m GPU. The subsequent processing consumes about 0.2s on CPU.

\subsection{Generalization Ability of Proposed Algorithm}

\begin{figure}[!tp]
\centering
\includegraphics[width=1.0\linewidth]{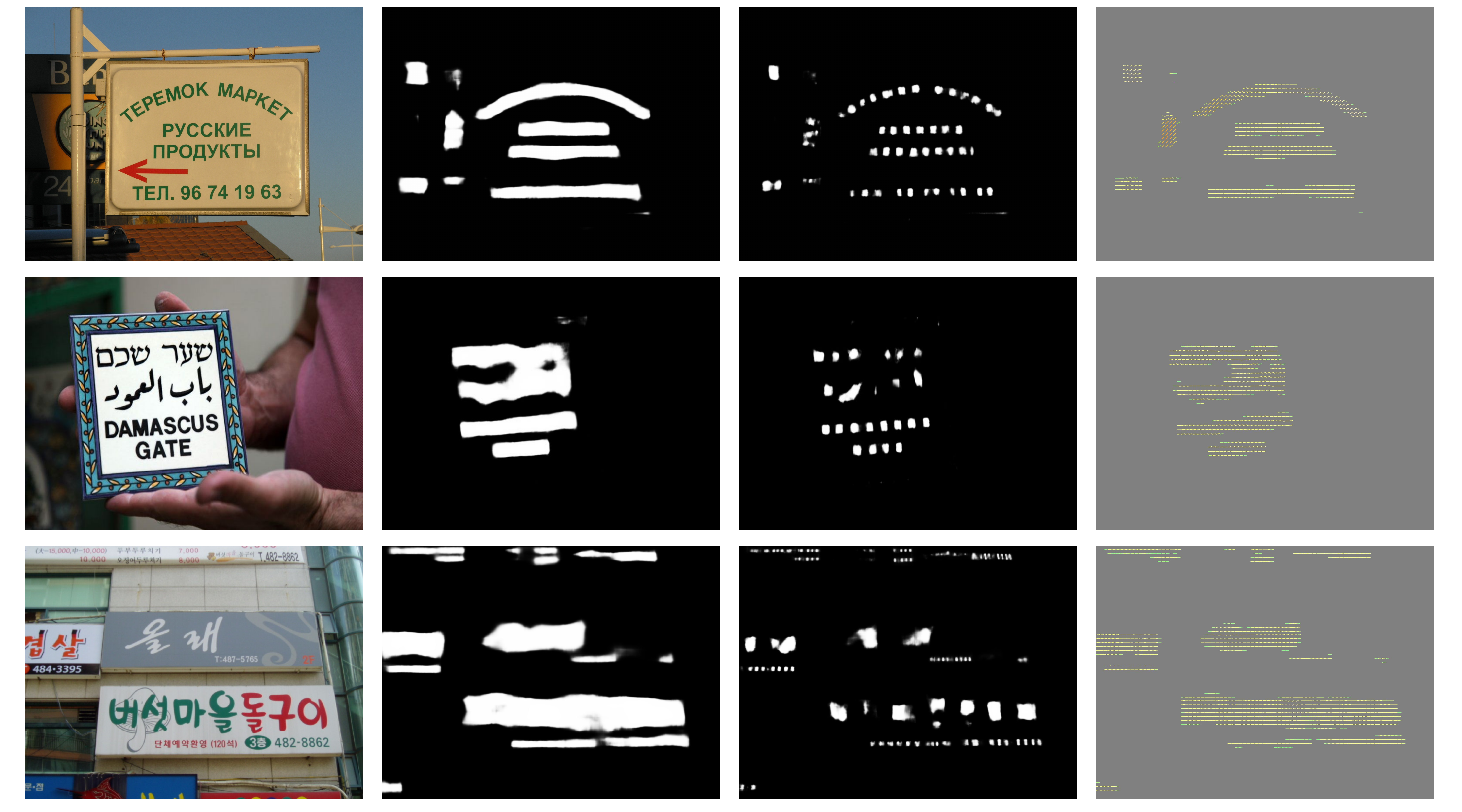}
\caption{Generalization to text of different scripts. Even though the algorithm is only trained on examples of English and Chinese, it can generalize well to different types of scripts, such as Arabic, Hebrew, Korean and Russian. Original image are harvested from the Internet.}   \label{Fig:Foreign}
\vspace{-2mm}
\end{figure}

\begin{figure}[!tp]
\centering
\includegraphics[width=0.98\linewidth]{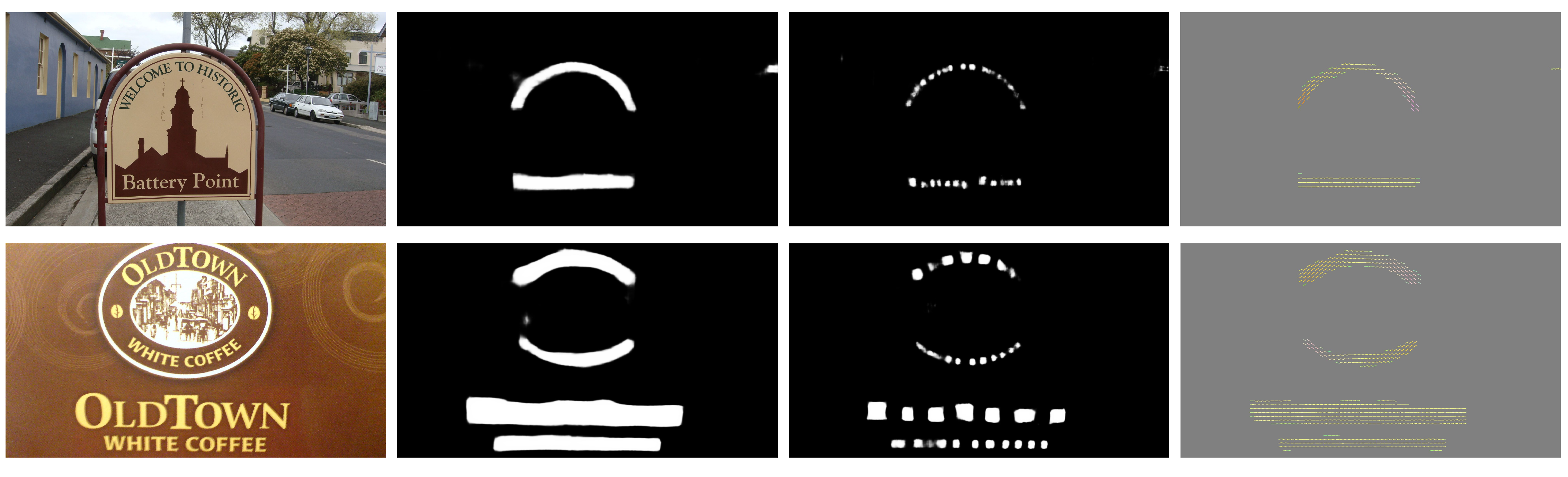}
\caption{Generalization to curved text. Curved examples are rarely seen in the training set, but the trained model can easily handle curved text. Original images are from~\cite{Ref:Risnumawan2014}.}  \label{Fig:Curved}
\vspace{-4mm}
\end{figure}

As can be seen from Fig.~\ref{Fig:Foreign} and Fig.~\ref{Fig:Curved}, although the proposed method was not trained with examples of various languages nor curved instances, it generalize well to such scenarios. This verifies the strong generalization ability of the proposed algorithm.

\subsection{Limitations of Proposed Algorithm}

The proposed method shows excellent capability in most scenarios, but it may fail in certain conditions. For example, it is found sensitive to serious blur and highlight. We believe if more examples and augmentation are introduced, these issues can be alleviated.

Moreover, the model size of the proposed method (about 56M) is still not compact enough and the speed without GPU (more than 14s for 640x480 images) is not sufficient, if we port the system to low-end PCs or mobile devices.

\section{Conclusions}  \label{Sec:Conclusions}

In this paper, we have presented a novel algorithm for text detection in natural scene images, which is based on the HED model~\cite{Ref:Xie2015}. The proposed algorithm is fundamentally different from majority of the previous methods in that: (1) It approaches scene text detection via whole image semantic segmentation, instead of local component extraction~\cite{Ref:Epshtein2010,  Ref:Neumann2011, Ref:Yao2012, Ref:Neumann2015} or window-based classification.~\cite{Ref:Wang2010, Ref:Neumann2013B, Ref:Jaderberg2014, Ref:Jaderberg2015B}, and thus is able to utilize more contextual information from larger scope. (2) It concurrently predicts the probability of text regions, characters and the linking orientation of nearby characters in a unified framework, while other methods~\cite{Ref:Epshtein2010, Ref:Neumann2011, Ref:Yao2012, Ref:Jaderberg2014, Ref:Yin2015} estimate such properties in separate stages. (3) It can directly spot multi-oriented and curved text from images, while most previous approaches~\cite{Ref:Yi2011, Ref:Bissacco2013, Ref:Jaderberg2015} have focused on horizontal or near-horizontal text. Moreover, the experiments on standard benchmark datasets demonstrate that the proposed method achieves superior performance than other competing systems.

Future directions worthy of further investigation and exploration might include: (1) Architecture. The 5-stage architecture of HED leads to breakthrough in edge detection, but this architecture is not necessarily the best for scene text detection. We would modify the network architecture and seek better ones that are suitable for this task. (2) Capacity. Training with more detailed labels (e.g., binary masks of character shapes) will endow the system with the ability to explicitly segment out character shapes from original images, which may potentially facilitate the subsequent text recognition procedure. (3) Efficiency. Adopting acceleration techniques~\cite{Ref:Jaderberg2014C} could largely speed up the proposed method, making it more practical.

\section{Acknowledgements}  \label{Sec:Acknowledgements}

{\small
\bibliographystyle{IEEEtran}
\bibliography{HolisticPrediction}
}

\end{document}